\documentclass{article}

\usepackage[final]{nips_2018}

\usepackage[utf8]{inputenc} 
\usepackage[T1]{fontenc}    
\usepackage{hyperref}       
\usepackage{url}            
\usepackage{booktabs}       
\usepackage{amsfonts}       
\usepackage{nicefrac}       
\usepackage{microtype}      
\usepackage{enumitem}
\usepackage{multicol}

\setcitestyle{numbers,open={[},close={]}}

\title{Some Requests for Machine Learning Research from the East African Tech Scene}

\author{
  Milan Cvitkovic\\
  Department of Computing and Mathematical Sciences\\
  California Institute of Technology\\
  Pasadena, CA 91106 \\
  \texttt{mcvitkov@caltech.edu} \\
}

\begin{document}

\maketitle

Based on 46 in--depth interviews with scientists, engineers, and CEOs, this document presents a list of concrete machine research problems, progress on which would directly benefit tech ventures in East Africa.\footnote{This work's focus on East Africa is based on where the author had work experience and connections.  But most of the problems listed are probably relevant to low--wealth--per--capita regions anywhere.}

The goal of this work is to give machine learning researchers a fuller picture of where and how their efforts as scientists can be useful.  The goal is thus \emph{not} to highlight research problems that are \emph{unique} to East Africa --- indeed many of the problems listed below are of general interest in machine learning.  The problems on the list are united solely by the fact that technology practitioners and organizations in East Africa reported a pressing need for their solution.

The author is aware that listing machine learning problems without also providing data for them is not a recipe for getting those problems solved.  If the reader is interested in any of the problems below, please get in touch.  I will gladly introduce them to the organizations or people with access to data for those problems.  But to protect privacy and intellectual property, I have not attributed problems to specific organizations or people in this document.

\section*{Research Problems}

\subsection*{Natural Language Processing}
Mobile phone ownership and use, particularly of feature phones, is widespread in East Africa.  SMS and voice interactions are one of the few big data sources in the region. 
Moreover, since literacy (technological and otherwise) remains low,  natural language interfaces and conversational agents have huge potential for impact.  

A few organizations in East Africa are trying to leverage NLP methods, but they face many challenges, due in part to the following.

\textbf{Handling Rapid Code--Switching with Models trained on Single Language Corpora -}
In SMS and voice communication, many East Africans rapidly code--switch (switch between languages).  This is usually done multiple times per sentence, throughout an interaction, and usually between English and another language.  

Despite perhaps striking some readers as a fringe linguistic phenomenon, every engineer interviewed herein who had worked with NLP models reported that this was a significant issue for them.  It makes using models trained on single--language corpora --- the most widely available corpora --- difficult.

\textbf{Named Entity Recognition with Multiple--Use Words -} NER is an important part of NLP pipelines in East Africa.  However, the entity detection step of NER is complicated by the fact that English words are commonly used as names in East Africa.  E.g. Hope, Wednesday, Silver, Editor, Angel, Constant. Capitalization is not used regularly enough in SMS to help.

\textbf{Location Extraction from Natural Language -}
Despite the proliferation of mobile phones, GPS availability and accuracy is limited in East Africa.  Extraction of locations from natural language is therefore critical for numerous applications, from localization of survey respondents to building speech--only navigation apps (for use with feature phones).

This task is complicated by the fact that most rural locales, and many urban ones, lack usable addressing schemes.  Most people specify locations and directions contextually, e.g.\ ``My house is in Kasenge; it's the yellow one two minutes down the dirt road across from the Medical Center.''  

Even approximate or probabilistic localization based on such location information from natural language would be invaluable.  Combining satellite imagery or user interaction would be particularly impressive.

\textbf{Priors for Autocorrect and Low Literacy SMS Use -}
SMS text contains many language misuses due to a combination of autocorrection and low literacy.  E.g.\ ``poultry farmer'' becoming ``poetry farmer''.  Such mistakes are bound to occur in any written language corpus, but engineers working with rural populations in East Africa report that this is a prevalent issue for them, confounding the use of pretrained language models.  This problem also exists to some degree in voice data with respect to English spoken in different accents.  Priors over autocorrect substitution rules, or custom, per--dialect confusion matrices between phonetically similar words could potentially help.

\textbf{Disambiguating Similar Languages -}
Numerous languages are spoken in East Africa, many of which are quite distinct in meaning but can be difficult to identify in small inputs or when rendered into text (especially when combined with typographical errors).  Even when data are available for tasks like sentiment analysis in multiple similar languages, performing tasks when the input language is ambiguous and from a set of similar languages remains an open problem.

\textbf{Data Gaps -}
Specific domains for which data and pretrained models are limited include East African languages; non--Latin--character text; non--Western English (East African English is fairly idiomatic).

\subsection*{Computer Vision}
Satellite data, mobile phone data (including mobile money), and in--depth, large--scale surveys or censuses (which are extremely rare) are the primary sources of big data in East Africa.  But among them, satellite data is the most abundant and open, which has led to widespread use of computer vision models for satellite data in the region.  Mobile phone cameras also have enabled the use of computer vision for applications ranging from disease identification to stock management.  Yet many research problems remain to be solved to maximize the utility of computer vision in East Africa.

\textbf{Specialized Models for Satellite Imagery\footnote{A more detailed explanation of this topic, written with the help of Dr.\ Hamed Alemohammad, can be found at \url{https://milan.cvitkovic.net/assets/documents/Satellite_Imagery_2018.pdf}.} -}
Excellent work has been done using satellite data in East Africa, and its importance to the region cannot be overstated.  But satellite data, as a subset of general image data, have many unique properties.  Little work has been done to develop specialized image models to exploit/compensate for these properties, some of which include:
\begin{itemize}[itemsep=0mm, leftmargin=5.5mm]
    \item The presence of reliable image metadata, such as precise geolocation of images on the Earth's surface, camera position and orientation, image acquisition time, and pixel resolution.
    \item The inherent time--series nature of satellite imagery, which consists of repeated images of the same location across time.
    \item Imagery that is captured at different wavelengths and modes, each with unique properties, e.g.\ optical vs.\ near--infrared or passive vs.\ active/radar.
    \item The presence of cloud occlusion (particularly in passive measurements), cloud shadows, and the ensuing illumination variability these both cause.
    \item Multiple resolutions of imagery for the same ground truth, and the frequent need to transfer models trained on one resolution to another.
\end{itemize}

\textbf{Map Generation -}
Generating road maps and identifying homes in rural areas are critical tasks for many East African organizations.  But road networks and buildings change rapidly in East Africa due to construction and weather, and road identification is challenging when roads are unpaved.  
Rapid, frequent, satellite--imagery--based map making is thus a high value use of computer vision in East Africa.

While the general task of generating maps and road--networks from satellite images is not new \cite{bastani_roadtracer:_2018}, the scarcity of labeled map data in East Africa means structured prediction models for map making in the region need to be developed that can  better leverage prior knowledge about road network structure.

\textbf{Document Understanding and OCR -}
Many East African government agencies, organizations, and businesses have all their records on paper.  Given the general scarcity of data in the region, solutions for automated information extraction from such documents is greatly desired.  These documents are usually handwritten, however, so existing OCR extraction pipelines have not proven usable.

\textbf{Data Gaps -}
Specific domains for which data and pretrained models are limited include: dark--skinned faces; high--resolution satellite imagery of East African geography; general ground--level imagery of East Africa (models pretrained on, e.g., ImageNet have trouble identifying East African consumer goods, vehicles, buildings, flora, etc.); low--resolution imagery of documents (bank statements, government records, IDs); images taken with feature phone cameras.

\subsection*{Data \emph{Au Naturel}}
Data from East Africa are scarce and expensive to obtain.  And they almost never come to practitioners as perfectly i.i.d.\ real--valued vector pairs.  Research into models better suited for data--as--they--are was the most common request heard from interviewees.

\textbf{Learning Directly on Relational Databases -}
Many data in East Africa are stored in relational, usually SQL, databases, including perhaps the most abundant, though least open, source of data in the region: private business data.   Extracting data from a relational database and converting them to real--valued vectors is one of the largest time expenditures of working data scientists and engineers in East Africa.  Moreover, converting the inherently relational data in a SQL database into vectors that a machine learning model can pretend are i.i.d.\ is, even when done by the best data scientist, a fraught process.

A system that could build a structured model, like a Recursive Neural Network, based on a relational database's particular schema and train it without requiring manual ETL and flattening of data into vectors would be beneficial for many of the organizations interviewed.

\textbf{Merging datasets -}
The most common way interviewees handled data scarcity was by merging datasets.  Versions of this tactic included:
\begin{itemize}[itemsep=0mm, leftmargin=5.5mm]
    \item Combining surveys containing differently--phrased versions of essentially the same question.
    \item Combining customer or surveyee interaction histories gathered over different but overlapping times.
    \item Augmenting survey data with satellite data, without accurate location information in the surveys.
\end{itemize}
No interviewee was familiar with any techniques or models well--suited for these scenarios.

\textbf{Adversarial Inputs (no, not that kind) -}
East African economies are low--trust relative to those of wealthier regions \cite{zak_trust_nodate}.  Interviewees who use machine learning with surveys or customer interaction data reported spending significant effort fighting fraud or dishonesty.

This issue is pervasive not only with companies that offer products, e.g.\ loans, based on survey responses.  It is present even in surveys where the surveyee stands to gain nothing.  Prof. Tim Brown suggests distinguishing such adversarial inputs into two categories:  ``defensive'', where surveyees do not trust your intent and thus obfuscate or misrepresent themselves in responses, and ``offensive'', where surveyees are searching for the answer you want to hear.

Handling fraud is not exclusively a machine learning problem by any means.  But there are interesting open research questions around building models, especially interactive systems, that are wary of being gamed.  E.g.\ safe RL models for when the danger is not extreme (negative) variance in rewards, but rather adversarial corruption of observations.

\subsection*{Resource--Limited Machine Learning}
Cellular access is widespread in East Africa, but it remains patchy in many rural areas, and electrical power for devices is scarce.  Additionally, some countries like Kenya have laws that prevent some data from leaving the country.  These issues limit the utility of existing machine learning technologies.

\textbf{Models for small, low--powered devices -}
The utility of reducing the computation and memory requirements of modern ML models is well known \cite{cheng_survey_2017}.  It is listed here simply to reiterate its importance to East African organizations trying to use deep learning.

\textbf{Communication--Limited Active Learning and Decision Making -}
Survey collection is an expensive, slow process, usually done by sending enumerators (human data collectors) across large regions to conduct interviews.  But it is also one of the only sources of data about East Africa, especially about poorer and rural regions. Maximizing the information collected in such surveys is thus critical.

One potential strategy to do this is active learning.  These days enumerators typically have access to smartphones with data--collection software like \href{https://opendatakit.org/software/}{ODK}, making it potentially viable to employ machine learning techniques like active learning in surveying.  However, this active learning scenario does not fit neatly into query synthesis, stream-based, or pool-based.  It is a multi--agent, cost--sensitive active learning task, where the agents cannot reliably communicate with one another, and where the agent's decision is not just whom to survey but also which subset of a large set of questions to ask each surveyee.

In a similar vein, some organizations are piloting automated money or food distribution programs in rural areas based on satellite, weather, survey, and other data.  This is a distributed decision--making task with the same complications as the surveying case described above.

\subsection*{Other}

\textbf{Reinforcement Learning -}
No interviewee reported using any reinforcement learning methods.  However, interest was expressed in it, particularly regarding machine teaching and using RL in simulations, e.g.\ using RL in epidemiological simulations to find worst case scenarios in outbreak planning.

\textbf{Machine Teaching -}
There is a shortage of good educational resources and teachers in East Africa.  Several initiatives exist that use mobile phones as an education platform.  Practitioners were interested in using ideas from machine teaching in their work to personalize content delivered.  However, the author did not encounter anyone who had employed any results from the machine teaching literature at this point.

\textbf{Uncertainty Quantification -}
An important factor that keeps the wealth of rich regions from moving into poorer regions like East Africa, despite the fact that it should earn greater returns there, is risk \cite{alfaro_why_2008}.  Not all risk can be machine--learned away by any means.  But (accurate) predictive models are risk--reduction tools.

Machine learning models are most useful for risk--reduction when they can (accurately) quantify their uncertainty.  This is particularly true when data are scarce, as they usually are in East Africa.  UQ is not a new problem by any means, but it is listed here to reiterate its importance to the organizations  interviewed.  Importantly, when used in East Africa, UQ is typically much more concerned with conservatively quantifying overall downside risk (with respect to some quantity of interest) than characterizing overall model uncertainty around point predictions.

\section*{Interviews Conducted}
\begin{multicols}{2}
\begin{itemize}[itemsep=0mm,leftmargin=5.5mm]
\item Chris Albon, Devoted Health
\item Hamed Alemohammad, Radiant Earth Foundation
\item Elvis Bando, Independent Data Science Consultant
\item Joanna Bichsel, Kasha
\item Prof. Tim Brown, Carnegie Mellon University Africa
\item Ben Cline, Apollo Agriculture
\item Johannes Ebert, Gravity.Earth
\item Dylan Fried, Lendable
\item Sam Floy, The East Africa Business Podcast
\item Lukas Lukoschek, MeshPower
\item Daniel Maison, Sky.Garden
\item Lauren Nkuranga, GET IT
\item Mehdi Oulmakki, African Leadership University
\item Jim Savage, Lendable
\item Rob Stanley, Wefarm
\item Linda Dounia Rebeiz, Eneza Education
\item Kamande Wambui, mSurvey
\item Muthoni Wanyoike, Instadeep
\item Anonymous individuals with the following affiliations: \\ Fenix International, Kasha, GET IT, Carnegie Mellon University Africa, kLab, Safaricom, Give Directly (2), Sankofa.africa, Rwanda Online, We Effect, Andela, Moringa School (2), Nelson Mandela African Institute of Science and Technology, IBM Research Africa, Sky.Garden (2), Medic Mobile, Engineering and data science students (9), independent data science consultant, anonymous company
\end{itemize}
\end{multicols}

\bibliography{references}

\begin{thebibliography}{4}
\providecommand{\natexlab}[1]{#1}
\providecommand{\url}[1]{\texttt{#1}}
\expandafter\ifx\csname urlstyle\endcsname\relax
  \providecommand{\doi}[1]{doi: #1}\else
  \providecommand{\doi}{doi: \begingroup \urlstyle{rm}\Url}\fi

\bibitem[Alfaro et~al.(2008)Alfaro, Kalemli-Ozcan, and
  Volosovych]{alfaro_why_2008}
Laura Alfaro, Sebnem Kalemli-Ozcan, and Vadym Volosovych.
\newblock Why {Doesn}'t {Capital} {Flow} from {Rich} to {Poor} {Countries}?
  {An} {Empirical} {Investigation}.
\newblock \emph{The Review of Economics and Statistics}, 90\penalty0
  (2):\penalty0 347--368, 2008.
\newblock \doi{10.1162/rest.90.2.347}.
\newblock URL \url{https://doi.org/10.1162/rest.90.2.347}.

\bibitem[Bastani et~al.(2018)Bastani, He, Abbar, Alizadeh, Balakrishnan,
  Chawla, Madden, and DeWitt]{bastani_roadtracer:_2018}
Favyen Bastani, Songtao He, Sofiane Abbar, Mohammad Alizadeh, Hari
  Balakrishnan, Sanjay Chawla, Sam Madden, and David~J. DeWitt.
\newblock {RoadTracer}: {Automatic} {Extraction} of {Road} {Networks} from
  {Aerial} {Images}.
\newblock 2018.

\bibitem[Cheng et~al.(2017)Cheng, Wang, Zhou, and Zhang]{cheng_survey_2017}
Yu~Cheng, Duo Wang, Pan Zhou, and Tao Zhang.
\newblock A {Survey} of {Model} {Compression} and {Acceleration} for {Deep}
  {Neural} {Networks}.
\newblock \emph{CoRR}, abs/1710.09282, 2017.
\newblock URL \url{http://arxiv.org/abs/1710.09282}.

\bibitem[Zak and Knack()]{zak_trust_nodate}
Paul~J. Zak and Stephen Knack.
\newblock Trust and {Growth}.
\newblock \emph{The Economic Journal}, 111\penalty0 (470):\penalty0 295--321.
\newblock \doi{10.1111/1468-0297.00609}.
\newblock URL
  \url{https://onlinelibrary.wiley.com/doi/abs/10.1111/1468-0297.00609}.

\end{thebibliography}

\end{document}